%% file: main.tex
\documentclass{article}
\usepackage{spconf,amsmath,graphicx,booktabs,url}
\usepackage{framed}

\usepackage{newtxmath}   
\usepackage{xcolor}
\colorlet{shadecolor}{gray!12}
\title{Benchmarking Vision-Language Models on Synapse and Neuron Proofreading in Connectomics}
\name{\begin{tabular}{@{}c@{}}
  Yicong Li$^{1,*}$, Junjie Wang$^{2}$, Leander Lauenburg$^{1}$, Ella Hugie$^{1}$ \\
  Alexandra Irger$^{1}$, Wanhua Li$^{3}$, Donglai Wei$^{2}$, Hanspeter Pfister$^{1,*}$\thanks{$^{*}$Corresponding authors: Yicong Li (yicong.li.0225@gmail.com) and Hanspeter Pfister (pfister@seas.harvard.edu).}
\end{tabular}}
\address{$^{1}$Harvard University, Cambridge, MA, USA \quad $^{2}$Boston College, Chestnut Hill, MA, USA \\ $^{3}$Nanyang Technological University, Singapore}

\begin{document}
\ninept   
\maketitle

\begin{abstract}
We benchmarked vision-language models (VLMs) on the decisions annotators take when inspecting electron microscopy images in connectomics: synapse proofreading (presence and polarity) and neuron proofreading (split errors and merge errors). For synapse proofreading, we evaluated 19 open and 2 closed models across various architectures and sizes under zero-shot, four-shot in-context learning and LoRA settings, against specialist models, on datasets constructed by us using public resources. For neuron proofreading, we evaluated 3 open and 2 closed models on the ConnectomeBench2 dataset, with cross-species transfer from fly and mouse to human and zebrafish. Most models were at chance zero-shot; a few examples helped mainly the closed and largest open ones. LoRA on a few thousand labels brought open models level with specialist models. When evaluated on unseen species, the best adapted VLMs outperformed specialist models in identifying merge errors. The project will be publicly available upon acceptance.
\end{abstract}


\section{Introduction}
\label{sec:intro}
Connectomics aims to reconstruct the wiring diagram of the brain: every neuron and every synapse between them. Currently, electron microscopy (EM), while expensive \cite{li2023xray2em,kuan2020dense}, is the primary imaging method that resolves both, at nanometer resolution over volumes that now reach peta-byte scale \cite{dorkenwald2024flywire,microns2025}, and the field has drawn considerable attention as these datasets appeared \cite{natmethods2025moty}. With the development of computer vision techniques, the reconstruction pipeline is mostly automatic in its two main stages: segmentation traces each neuron through the volume \cite{funke2019,lee2017superhuman}, and synapse detection finds the contacts between them and assigns their direction \cite{heinrich2018synaptic,buhmann2021synapses}. However, both stages still need human eyes. Synapse detectors are validated and corrected by annotators who judge, section by section, whether a contact is a synapse and which side is presynaptic. Segmentations are proofread, because a split or merge error propagates through every circuit the neuron takes part in; FlyWire took about three million proofreading edits \cite{dorkenwald2024flywire,schlegel2024}, and MICrONS proofreading continues years after imaging \cite{microns2025}. Each of these checks is a small visual decision taken on a few EM sections; we call them synapse and neuron proofreading.

Vision-language models (VLMs) answer visual questions in natural language and can be steered by a few examples or a short fine-tuning run, yet on general microscopy benchmarks they still struggle even with modality identification \cite{mubench}. Whether they can make an annotator's decision from raw EM, and what it takes to get them there, is largely unknown.

Prior work on these two tasks is supervised and task-specific. Synapse
detection has been posed as voxel classification \cite{heinrich2018synaptic},
as partner assignment from point annotations \cite{buhmann2021synapses,li2024waspsyn}, and as
direct mask generation for the pre- and postsynaptic partners of a given cleft
\cite{turner2020synaptic}. On the proofreading side, guided
proofreading trains CNNs to flag split and merge errors and to propose the
correction \cite{haehn2018guided}. More recently, ConnectomeBench
\cite{connectomebench} evaluated several multimodal LLMs on proofreading tasks, but only on mesh renderings rather than EM images. Its successor, ConnectomeBench2 \cite{connectomebench2} released expert-labelled sites with EM slices from four species and trained a ViT to compare with human performance, without evaluating language models.

To the best of our knowledge, no prior study has evaluated a range of VLMs on
synapse detection from EM, and none has benchmarked VLMs on neuron proofreading using ConnectomeBench2
\cite{connectomebench2}, whose only published results are for a supervised ViT. Concretely, our contributions are:

\begin{itemize}\itemsep1pt \parskip0pt \topsep2pt
\item We built a synapse proofreading benchmark from public fly and mouse connectomes and evaluated 19 open and two closed VLMs on it zero-shot, with four in-context examples, and with LoRA, against specialist models.
\item We presented the first evaluation of VLMs on ConnectomeBench2: three open models were run on it in full against the dataset's own ViT, and the same models were compared with two closed VLMs and three specialist networks on mouse and fly subsets.
\item We found that most open models were at chance zero-shot, many returning one fixed answer, while the closed models performed above chance on neuron proofreading. In-context examples helped the closed models and the largest open ones, and LoRA fine-tuning brought open models level with the specialists. Additionally, VLMs seem to be more resilient to cross-species transfer compared with specialist models.
\end{itemize}

\section{Benchmark}
\label{sec:bench}

\subsection{Task 1: Synapse proofreading}
\label{sec:t1}
We built the synapse proofreading tasks from two public sources with ground-truth synapse annotations, CREMI \cite{cremi} for fly and MICrONS \cite{microns2025} for mouse. Every item is a single EM section of $256\times256$ pixels at 8\,nm per pixel, a $2.05\,\mu$m field of view. For the \emph{presence} subtask, a yellow circle of 224\,nm radius marks the query point and the model is asked whether a chemical synapse is present there. For the \emph{polarity} subtask, two markers, A and B, are placed inside the two partner processes of a synapse, with the letters assigned at random, and the model is asked which marker is presynaptic.

\begin{figure}[t]
\centering
\includegraphics[width=0.7\columnwidth]{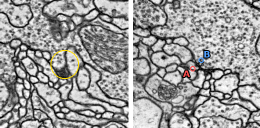}
\caption{Synapse proofreading example inputs of fly. Left: \emph{presence}, the query point marked by the yellow circle (answer: yes). Right: \emph{polarity}, markers A and B inside the two partner processes (answer: B).}
\label{fig:t1}
\end{figure}

Fig.~\ref{fig:t1} shows one item of each subtask and below is the accompanying prompt:

\begin{shaded*}\noindent
\textbf{Preamble (both subtasks):} This is a 2 um x 2 um electron micrograph of brain tissue (adult Drosophila brain), imaged at 8 nm per pixel. Dark membranes outline neuronal processes; mitochondria appear as dark oval organelles; synaptic vesicles are small round $\sim$40 nm blobs clustered inside axon terminals.
\textbf{Presence:} A yellow circle marks the location of interest. Is there a chemical synapse (a synaptic cleft with a presynaptic vesicle cluster on one side and a postsynaptic density on the other) at the marked location? Answer with exactly one word: yes or no.
\textbf{Polarity:} Two markers, A and B, are placed inside two adjacent neuronal processes that form a chemical synapse at their shared membrane. Which marker is inside the PRESYNAPTIC process, i.e. the side that contains the cluster of synaptic vesicles? Answer with exactly one letter: A or B.
\end{shaded*}

For fly we used the three CREMI volumes \cite{cremi}, adult \emph{Drosophila} brain at $4\times4\times40$\,nm with hand-labelled synaptic clefts and their pre- and postsynaptic partners. We assigned volume~A to training, B to validation and C to testing, so the three splits come from different brain regions. Positive samples for the \emph{presence} subtask were created by placing one query at the centroid of every cleft, in the section where the cleft is largest. For each positive we then drew one negative at a random location that lies at least 256\,nm from any cleft voxel in its section and 600\,nm from every annotated synapse, and we kept the query point at the same offset from the crop centre as in the paired positive, so that the offset alone carries no information. For \emph{polarity} we took the annotated partner pairs and placed the markers 72 to 240\,nm from the cleft and at least 112\,nm apart. This procedure yielded 246, 262 and 330 \emph{presence} items for training, validation and test, half of them positive, and 170, 275 and 413 \emph{polarity} pairs.

For mouse we used the public MICrONS cubic-millimetre volume of visual cortex (minnie65) \cite{microns2025}, available at $8\times8\times40$\,nm in its public mirror, which comes with a cleft segmentation and an automatically generated synapse table. Because the table is automatic, we kept only entries of at least 300 voxels whose cleft is confirmed at the crop centre, and we drew negatives at random points with no cleft within 256\,nm. To keep the splits apart, we divided the volume into contiguous blocks along one axis, separated by gaps, so that training, validation and test tissue never overlap. This gave 6000, 1000 and 3000 samples for the \emph{presence} subtask, half positive, and 3000, 500 and 1500 \emph{polarity} pairs.

\subsection{Task 2: Neuron proofreading}
\label{sec:t2}

\begin{table*}[t]
\centering\small\setlength{\tabcolsep}{3.4pt}\renewcommand{\arraystretch}{0.85}
\caption{Synapse proofreading, balanced accuracy (\%) on the test splits. P: \emph{presence}; Pol: \emph{polarity}. Bold: best VLM per column; \underline{underlined}: second best; \textit{Italic rows}: specialist models, not ranked; --: not applicable.}
\label{tab:t1}
\input{tables/tab_t1}
\end{table*}

For neuron proofreading we did not build a new dataset but adopted ConnectomeBench2 (CB2) \cite{connectomebench2}, which released proofreading sites from the edit histories of expert proofreaders, together with matched controls, in mouse, fly, human and zebrafish. Each site comes as four EM views taken before the edit, the $xy$, $xz$ and $yz$ planes and an oblique plane, at $224\times224$ pixels, together with the masks of the segments involved; the cutouts span $2.5\,\mu$m in mouse and human and $1.5\,\mu$m in fly and zebrafish. We kept the released images unchanged and only tinted the masks, so that the model knows which segments the question is about. In the \emph{split-error} subtask, the two segments are tinted red and blue in all four planes, and the answer is yes when the proofreaders merged them, that is, at a split error. In the \emph{merge-error} subtask, the union of the segments is tinted red in the three planes, and the answer is no when the proofreaders split it, that is, at a merge error. We dropped the oblique plane for this task because its orientation is chosen from the post-split masks and would leak the answer, as the dataset card itself warns.

\begin{figure}[t]
\centering
\includegraphics[width=0.9\columnwidth]{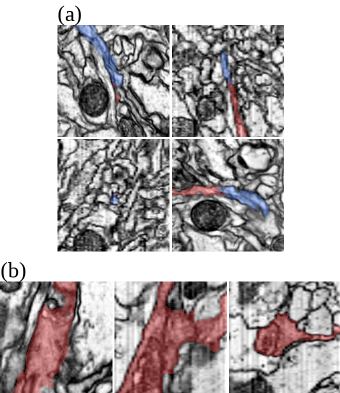}
\caption{Neuron proofreading example inputs of fly. (a) A \emph{split-error} site: the two segments are tinted red and blue; the proofreaders merged them, so the answer is yes. (b) A \emph{merge-error} site: the segment is tinted red; the proofreaders split it, so the answer is no.}
\label{fig:t2}
\end{figure}

Fig.~\ref{fig:t2} shows one item of each subtask and below is the accompanying prompt:

\begin{shaded*}\noindent
\textbf{Split error:} The four panels are electron-microscopy slices through the same location in adult Drosophila brain, each about 1.5 um across: top-left xy, top-right xz, bottom-left yz, bottom-right an oblique plane. Two segments from an automatic neuron segmentation are tinted RED and BLUE and meet near the centre. Automatic segmentations sometimes wrongly split one neuron into two segments, whereas two different cells are separated by a continuous dark membrane. Do the red and blue segments belong to the SAME neuron, so that they should be merged? Answer with exactly one word: yes or no.
\textbf{Merge error:} The three panels are electron-microscopy slices (xy, xz, yz) through the same location in adult Drosophila brain, each about 1.5 um across. One segment from an automatic neuron segmentation is tinted RED. Automatic segmentations sometimes wrongly merge two different cells across a membrane near the centre of the view. Is the red segment a SINGLE neuron here, with no merge error at the centre? Answer with exactly one word: yes or no.
\end{shaded*}

We evaluated neuron proofreading on three different setups. The first is the complete test split of the dataset's second release, 101\,929 operations with their natural class mix, namely 20\,879 split-error and 22\,094 merge-error operations in mouse, 24\,065 and 24\,203 in fly, 5\,351 and 60 in human, and 5\,223 and 54 in zebrafish; models were trained on its complete training and validation split of 531\,734 and 82\,822, respectively. The second is a balanced subset of mouse and fly with 3000, 600 and 1000 sites per species and task for training, validation and test, on which the models can be compared with closed-source VLMs and specialist networks under our budget. The third measures transfer to the species that no model developed using the second setup ever saw in training: namely, human and zebrafish.

\subsection{Evaluation protocol}
\label{sec:protocol}
We report balanced accuracy throughout and each model saw one image and one prompt per item; the prompt states the image scale, the marker convention and the definition of the target, and asks for exactly one word or letter. We decoded greedily with at most 64 new tokens and took the last \emph{yes}/\emph{no} or \emph{A}/\emph{B} in the reply as the decision; any other reply counts as wrong. All three regimes shared the same images and prompts. In the zero-shot regime, the model received only the test sample. In the 4-shot regime, we prepended four training examples of the same species and subtask as prior turns, two for each answer. In the LoRA regime, one adapter per task was fine-tuned on both subtasks and all training species with the recipe given in Sec.~\ref{sec:models}.

\subsection{Models}
\label{sec:models}
For synapse proofreading, we evaluated several open families of instruction-tuned VLMs across their size ladders: Qwen2.5-VL (3B, 7B, 32B) \cite{qwen25vl}, Qwen3-VL (2B, 4B, 8B, 32B) \cite{qwen3vl}, Qwen3.5 (2B, 4B, 9B, 27B) \cite{qwen35}, InternVL3 (2B, 8B, 14B, 38B) \cite{internvl3} and Gemma~4 (E2B, E4B, 12B, 31B) \cite{gemma4}. For neuron proofreading, due to budget constraints, we kept three mid-sized models, one each from the Qwen, InternVL, and Gemma lines: Qwen3.5-9B, InternVL3-8B and Gemma~4 12B. We also evaluated two closed models, GPT-5.6 Sol and Claude Opus 5, as zero-shot and in-context reference rows.

To fine-tune the open VLMs we used LoRA \cite{hu2022lora} with one recipe throughout: rank-16 adapters with $\alpha=32$ and dropout 0.05 on the attention and MLP projections of the language model, with the vision encoder frozen and the base weights frozen in bfloat16, AdamW at a learning rate of $10^{-4}$, a short linear warm-up followed by a cosine schedule, and only the answer token and the end-of-turn token supervised.

As references, we fine-tuned three types of specialist models: ResNet-50 \cite{he2016resnet} and ConvNeXt-Tiny \cite{liu2022convnext}, initialized from ImageNet, and a ViT-S/14 initialized from DINOv2 \cite{oquab2023dinov2}.

\section{Results}
\label{sec:results}

\begin{table}[t]
\centering\small\setlength{\tabcolsep}{3pt}\renewcommand{\arraystretch}{0.85}
\caption{Neuron proofreading on the complete ConnectomeBench2, balanced accuracy (\%) on the test split: all 101,929 samples across 4 species. S: \emph{split-error} subtask, M: \emph{merge-error} subtask. Bold: best VLM per column; \textit{Italic row}: the dataset's own reported EM-only ViT-B; --: not applicable.}
\label{tab:t2}
\input{tables/tab_t2-1}
\end{table}

\begin{table}[t]
\centering\small\setlength{\tabcolsep}{3.4pt}\renewcommand{\arraystretch}{0.85}
\caption{Neuron proofreading on the subset of ConnectomeBench2 (Sec.~\ref{sec:t2}), balanced accuracy (\%). S: \emph{split-error} subtask, M: \emph{merge-error} subtask. Bold: best VLM per row; \textit{italic columns}: specialist models, not ranked; --: not applicable.}
\label{tab:t2-2}
\input{tables/tab_t2-2}
\end{table}

\subsection{Synapse proofreading}
\label{sec:res-t1}

Table~\ref{tab:t1} shows the synapse proofreading results. Zero-shot, most open models could not interpret the images: across the 76 cells of the 19 open models the median score was 50.0 and 62 cells fell below 55. The exceptions were the two largest Gemma~4 and Qwen3.5 models, mostly on fly, where Gemma~4 31B reached 73.6 on \emph{presence} and Gemma~4 12B 75.0 on \emph{polarity}; on mouse no open model passed 67. The closed models stayed at or near chance on mouse, but GPT-5.6 Sol read fly \emph{polarity} at 77.2, the highest zero-shot value in the table, while staying at 59.7 on fly \emph{presence}.

Four in-context examples raised the performance: Gemma~4 12B reached 83.2 on fly \emph{polarity}, above the best specialist at 82.6, Gemma~4 31B 81.8 on fly \emph{presence}, and Qwen3.5-9B 75.8 on fly \emph{presence}, whereas the best mouse cells stayed at 64.4 and 66.9. GPT-5.6 Sol reached 79.4 and 83.1 on the two fly subtasks, and Claude Opus 5 68.2 and 65.7.

LoRA changed the picture entirely. 63 of 76 cells reached at least 70, and the best adapter per cell reached 85.5 on fly \emph{presence}, 91.1 on fly \emph{polarity}, 93.5 on mouse \emph{presence} and 94.5 on mouse \emph{polarity}, ahead of or level with the specialists on fly and within three points of them on mouse. Size mattered less once adapted: Gemma~4 12B beat Gemma~4 31B on three of the four cells, and Qwen3.5-9B matched or beat Qwen3.5-27B on all four.

\subsection{Neuron proofreading}

\textbf{Full ConnectomeBench2 evaluation.} Table~\ref{tab:t2} reports the three open models evaluated on the complete ConnectomeBench2 (CB2) under the dataset's own protocol. Zero-shot, all three sat at or near chance. Four in-context examples lifted the \emph{split-error} subtask scores of Qwen3.5-9B and Gemma~4 12B to 58.7 and 62.7 and the \emph{merge-error} scores of all three to 57 to 58. LoRA brought all three level with the dataset's own EM-only ViT-B, at 94.2 and 91.1: Gemma~4 12B reached 95.0 and 90.9, and the other two were within about one point on both tasks.

\noindent\textbf{Balanced subsets, closed models, and transfer.} Table~\ref{tab:t2-2} compares the same three models with the two closed models and the specialist models on the balanced mouse and fly subset of CB2, and shows how their adapters transfer. Zero-shot, the open models were again at or near chance, apart from Qwen3.5-9B at 59.3 on mouse merge errors, whereas the closed models performed better: Claude Opus 5 scored 73.1 and 64.7 on split errors in mouse and fly, GPT-5.6 Sol 67.2 and 63.0. With four examples the closed models rose to 76.4 and 74.4 at best, Gemma~4 12B and Qwen3.5-9B reached 58 to 69, and InternVL3-8B barely moved. LoRA took the open models to 83.1 to 93.3 on split errors and 74.5 to 84.6 on merge errors, within 1.6 points of the references on mouse, 3.5 points behind on fly split errors, and ahead on fly merge errors, 80.1 against 79.7. Applied unchanged to human and zebrafish, the adapters kept most of this accuracy. On split errors the best adapter transferred about as well as the best specialist network, Qwen3.5-9B at 83.1 and 88.0 against 85.4 and 89.3 for ConvNeXt-Tiny. On merge errors the best adapter beat the best network on both species, Gemma~4 12B reaching 77.3 and 82.7 against 66.9 and 74.7, and on human all three adapters did.


\section{Discussion and conclusion}
\label{sec:conclusion}
We benchmarked vision-language models for synapse and neuron proofreading tasks on EM images in connectomics. Without adaptation, most open models are at or near chance and cannot be dropped into an annotation pipeline; the closed models performed better under zero-shot and gained from in-context examples, but stay well below the adapted open models. A few thousand labels with LoRA bring an open model level with a purpose-trained network, and the adapters carry to species never seen in training, the best of them beating the specialist networks on merge errors. What makes the VLM worth its cost is not accuracy in the trained setting, where a small network is about as good and cheaper per image, but what happens away from it. One adapter per task answered both questions in both species, where the specialist models needed one network per question, and the adapters generally lost less accuracy on unseen species, which suggests that large-scale pretraining on images and text supplies context the small models lack. A VLM is therefore the right tool when the decisions are many or when a volume comes from a species with no training data; a purpose-trained network remains the right tool for one fixed decision on one well-labelled dataset.


\clearpage

\section{Compliance with Ethical Standards}
This retrospective study used only publicly released data, CREMI, MICrONS and ConnectomeBench2, including its human sites from the public H01 release; no new human or animal data were collected and ethical approval was not required.

\section{Acknowledgments}
This work was supported by NSF grant NCS-FO-2124179 and NIH grant R01HD104969. The authors have no relevant financial or non-financial interests to disclose. We thank Jingpeng Wu for valuable discussions and feedback. AI tools were used only for editing and grammar enhancement of the manuscript.

\bibliographystyle{IEEEbib}
\bibliography{refs}

\end{document}

%% file: tables/tab_t1.tex
\begin{tabular}{@{}l cccc cccc cccc@{}}
\toprule
 & \multicolumn{4}{c}{zero-shot} & \multicolumn{4}{c}{4-shot} & \multicolumn{4}{c}{LoRA} \\
\cmidrule(lr){2-5}\cmidrule(lr){6-9}\cmidrule(lr){10-13}
Model & fly P & fly Pol & mouse P & mouse Pol & fly P & fly Pol & mouse P & mouse Pol & fly P & fly Pol & mouse P & mouse Pol \\
\midrule
Qwen2.5-VL-3B & 49.1 & 50.0 & 49.5 & 50.0 & 50.9 & 50.0 & 53.9 & 50.0 & 50.3 & 57.8 & 74.4 & 76.4 \\
Qwen2.5-VL-7B & 50.0 & 49.0 & 49.6 & 49.2 & 48.8 & 49.1 & 48.2 & 49.3 & 71.8 & 70.0 & 86.3 & 81.4 \\
Qwen2.5-VL-32B & 48.2 & 49.1 & 46.2 & 49.8 & 55.2 & 49.4 & 52.2 & 50.3 & 68.2 & 61.7 & 83.5 & 77.4 \\
\addlinespace[2pt]
Qwen3-VL-2B & 51.2 & 49.1 & 48.3 & 50.9 & 50.0 & 48.7 & 54.1 & 50.0 & 69.7 & 56.9 & 81.0 & 79.9 \\
Qwen3-VL-4B & 53.3 & 48.2 & 47.5 & 51.9 & 59.1 & 51.7 & 48.8 & 49.8 & 72.4 & 56.3 & 83.7 & 82.1 \\
Qwen3-VL-8B & 52.4 & 56.8 & 47.1 & 49.1 & 57.6 & 54.7 & 56.5 & 49.9 & 73.0 & 66.0 & 83.1 & 87.8 \\
Qwen3-VL-32B & 50.0 & 54.9 & 49.9 & 51.3 & 71.2 & 58.4 & 56.1 & 50.7 & 71.5 & \underline{76.2} & 82.9 & 89.2 \\
\addlinespace[2pt]
Qwen3.5-2B & 53.0 & 56.9 & 51.7 & 49.7 & 57.0 & 52.1 & 55.9 & 48.7 & 75.5 & 65.1 & 87.1 & 87.6 \\
Qwen3.5-4B & 50.0 & 50.5 & 50.0 & 50.1 & 52.1 & 64.2 & 51.0 & 50.3 & 80.0 & 74.6 & 87.9 & \underline{90.1} \\
Qwen3.5-9B & 50.0 & 71.9 & 50.0 & 52.5 & 75.8 & 72.4 & 56.5 & 49.4 & \underline{82.7} & 75.4 & 89.0 & 89.9 \\
Qwen3.5-27B & \underline{68.5} & 58.5 & 55.4 & 51.8 & 73.0 & 69.0 & \underline{63.9} & 52.4 & \underline{82.7} & 72.0 & 87.5 & 89.3 \\
\addlinespace[2pt]
InternVL3-2B & 49.7 & 50.2 & 49.3 & 50.1 & 50.0 & 48.0 & 49.9 & 46.0 & 77.3 & 61.2 & 88.8 & 77.2 \\
InternVL3-8B & 51.5 & 45.2 & 48.4 & 48.4 & 67.6 & 46.1 & 47.1 & 46.3 & 79.1 & 73.4 & 89.8 & 87.0 \\
InternVL3-14B & 51.5 & 47.1 & 50.4 & 48.4 & 60.3 & 48.4 & 52.5 & 48.2 & 80.0 & 72.3 & 89.8 & 86.9 \\
InternVL3-38B & 50.6 & 47.2 & 56.1 & 49.5 & 65.2 & 56.3 & 62.2 & 51.2 & 80.6 & 75.6 & \underline{91.6} & 88.6 \\
\addlinespace[2pt]
Gemma~4 E2B & 52.1 & 48.4 & 47.3 & 49.8 & 51.2 & 49.6 & 49.8 & 52.7 & 60.9 & 53.6 & 79.4 & 76.2 \\
Gemma~4 E4B & 50.0 & 52.1 & 49.5 & 50.2 & 65.5 & 59.0 & 48.1 & 48.2 & 72.1 & 59.0 & 83.2 & 75.5 \\
Gemma~4 12B & 60.6 & \underline{75.0} & 51.8 & \underline{61.5} & 72.1 & \textbf{83.2} & 56.4 & \underline{61.5} & 81.5 & \textbf{91.1} & \textbf{93.5} & \textbf{94.5} \\
Gemma~4 31B & \textbf{73.6} & 70.1 & \textbf{60.9} & \textbf{66.6} & \textbf{81.8} & 76.0 & \textbf{64.4} & \textbf{66.9} & \textbf{85.5} & 75.7 & 88.0 & 88.1 \\
\midrule
GPT-5.6 Sol & 59.7 & \textbf{77.2} & 45.0 & 56.8 & \underline{79.4} & \underline{83.1} & 56.3 & 60.9 & -- & -- & -- & -- \\
Claude Opus 5 & 59.7 & 63.4 & \underline{57.2} & 52.0 & 68.2 & 65.7 & 62.2 & 53.7 & -- & -- & -- & -- \\
\midrule
\textit{ResNet-50} & \textit{--} & \textit{--} & \textit{--} & \textit{--} & \textit{--} & \textit{--} & \textit{--} & \textit{--} & \textit{83.8} & \textit{75.9} & \textit{95.6} & \textit{95.3} \\
\textit{ConvNeXt-T} & \textit{--} & \textit{--} & \textit{--} & \textit{--} & \textit{--} & \textit{--} & \textit{--} & \textit{--} & \textit{85.6} & \textit{82.6} & \textit{96.4} & \textit{95.6} \\
\textit{ViT-S/14 (DINOv2)} & \textit{--} & \textit{--} & \textit{--} & \textit{--} & \textit{--} & \textit{--} & \textit{--} & \textit{--} & \textit{79.7} & \textit{60.5} & \textit{95.3} & \textit{92.2} \\
\bottomrule
\end{tabular}

%% file: tables/tab_t2-1.tex
\begin{tabular}{@{}l cc cc cc@{}}
\toprule
 & \multicolumn{2}{c}{zero-shot} & \multicolumn{2}{c}{4-shot} & \multicolumn{2}{c}{LoRA} \\
\cmidrule(lr){2-3}\cmidrule(lr){4-5}\cmidrule(lr){6-7}
Model & S & M & S & M & S & M \\
\midrule
Qwen3.5-9B & 50.0 & \textbf{52.9} & 58.7 & 57.3 & 94.6 & 90.2 \\
InternVL3-8B & \textbf{51.0} & 50.0 & 49.0 & 57.3 & 93.9 & 90.1 \\
Gemma~4 12B & 50.1 & 50.0 & \textbf{62.7} & \textbf{58.0} & \textbf{95.0} & \textbf{90.9} \\
\midrule
\textit{CB2 ViT-B, EM only} & \textit{--} & \textit{--} & \textit{--} & \textit{--} & \textit{94.2} & \textit{91.1} \\
\bottomrule
\end{tabular}

%% file: tables/tab_t2-2.tex
\begin{tabular}{@{}l ccc|cc|ccc@{}}
\toprule
 & \rotatebox{90}{Qwen3.5-9B} & \rotatebox{90}{InternVL3-8B} & \rotatebox{90}{Gemma~4 12B} & \rotatebox{90}{GPT-5.6 Sol} & \rotatebox{90}{Claude Opus 5} & \rotatebox{90}{\textit{ResNet-50}} & \rotatebox{90}{\textit{ConvNeXt-T}} & \rotatebox{90}{\textit{ViT-S/14}} \\
\midrule
\multicolumn{9}{@{}l}{\emph{zero-shot}} \\
\quad mouse S & 50.0 & 53.4 & 50.4 & 67.2 & \textbf{73.1} & \textit{--} & \textit{--} & \textit{--} \\
\quad mouse M & 59.3 & 50.0 & 50.0 & \textbf{67.6} & 67.3 & \textit{--} & \textit{--} & \textit{--} \\
\quad fly S & 50.0 & 52.0 & 50.0 & 63.0 & \textbf{64.7} & \textit{--} & \textit{--} & \textit{--} \\
\quad fly M & 51.0 & 50.0 & 50.0 & \textbf{57.0} & \textbf{57.0} & \textit{--} & \textit{--} & \textit{--} \\
\addlinespace[2pt]
\multicolumn{9}{@{}l}{\emph{4-shot}} \\
\quad mouse S & 60.2 & 49.7 & 64.5 & 74.1 & \textbf{76.4} & \textit{--} & \textit{--} & \textit{--} \\
\quad mouse M & \textbf{69.1} & 54.1 & 57.7 & 65.7 & 68.0 & \textit{--} & \textit{--} & \textit{--} \\
\quad fly S & 60.4 & 48.9 & 64.7 & \textbf{74.4} & 66.4 & \textit{--} & \textit{--} & \textit{--} \\
\quad fly M & 61.7 & 50.0 & 60.0 & \textbf{67.7} & 61.9 & \textit{--} & \textit{--} & \textit{--} \\
\addlinespace[2pt]
\multicolumn{9}{@{}l}{\emph{LoRA, trained on mouse + fly}} \\
\quad mouse S & \textbf{93.3} & 89.3 & 92.5 & -- & -- & \textit{94.9} & \textit{94.9} & \textit{85.3} \\
\quad mouse M & \textbf{84.6} & 84.4 & 83.4 & -- & -- & \textit{86.1} & \textit{85.9} & \textit{82.0} \\
\quad fly S & \textbf{86.6} & 83.1 & 85.2 & -- & -- & \textit{89.2} & \textit{90.1} & \textit{80.8} \\
\quad fly M & \textbf{80.1} & 78.8 & 74.5 & -- & -- & \textit{76.7} & \textit{79.7} & \textit{74.2} \\
\addlinespace[2pt]
\multicolumn{9}{@{}l}{\emph{transfer to unseen species}} \\
\quad human S & \textbf{83.1} & 72.2 & 79.4 & -- & -- & \textit{81.9} & \textit{85.4} & \textit{65.7} \\
\quad human M & 71.9 & 67.6 & \textbf{77.3} & -- & -- & \textit{59.8} & \textit{66.9} & \textit{57.9} \\
\quad zebrafish S & \textbf{88.0} & 79.0 & 85.4 & -- & -- & \textit{87.3} & \textit{89.3} & \textit{78.8} \\
\quad zebrafish M & 77.3 & 73.5 & \textbf{82.7} & -- & -- & \textit{70.5} & \textit{74.7} & \textit{65.6} \\
\bottomrule
\end{tabular}